\def\year{2019}\relax 
\documentclass[letterpaper]{article} 
\usepackage{aaai19}  
\usepackage[T1]{fontenc}    
\usepackage[utf8]{inputenc}
\usepackage{times}  
\usepackage{helvet}  
\usepackage{courier}  
\usepackage{hyperref}
\usepackage{url}  
\usepackage{booktabs}
\usepackage{amsfonts}
\usepackage{nicefrac}
\usepackage{microtype}
\usepackage{amsmath}
\usepackage{amssymb}
\usepackage{bbm}
\usepackage{multirow}
\usepackage{threeparttable}
\usepackage{algorithmicx,algorithm,eqparbox,array}
\algnewcommand{\LineComment}[1]{\State // #1}
\usepackage[noend]{algpseudocode}
\usepackage{graphicx}  
\usepackage[dvipsnames]{xcolor}
\usepackage{color, colortbl}
\usepackage{subfigure}
\usepackage{tikz}

\usetikzlibrary{matrix,arrows,decorations.pathmorphing}
\tikzset{node style ge/.style={rectangle}}
\graphicspath{{figures/}}

\nocopyright

\begin{document}

\title{Learning to Solve NP-Complete Problems: A Graph Neural Network for Decision TSP}

\author{
  Marcelo Prates\thanks{Equal contribution} \\
  Institute of Informatics\\
  UFRGS\\
  Porto Alegre, Brazil \\
  {morprates@inf.ufrgs.br} \\
  \And
  Pedro Avelar{$^*$} \\
  Institute of Informatics\\
  UFRGS\\
  Porto Alegre, Brazil \\
  {pedro.avelar@inf.ufrgs.br} \\
  \And
  Henrique Lemos{$^*$} \\
  Institute of Informatics\\
  UFRGS\\
  Porto Alegre, Brazil \\
  {hlsantos@inf.ufrgs.br} \\
  \And
  Luis C. Lamb \\
  Institute of Informatics\\
  UFRGS\\
  Porto Alegre, Brazil \\
  {lamb@inf.ufrgs.br} \\
  \And
  Moshe Y. Vardi \\
  Dept. of Computer Science \\
  Rice University \\
  Houston, USA \\
  {vardi@cs.rice.edu} \\
  }

\makeatletter
\def\BState{\State\hskip-\ALG@thistlm}
\makeatother

\algnewcommand{\LineComment}[1]{\State // #1}

\maketitle

\begin{abstract}
Graph Neural Networks (GNN) are a promising technique for bridging differential programming and combinatorial domains. GNNs employ trainable modules which can be assembled in different configurations that reflect the relational structure of each problem instance. 
In this paper, we show that GNNs can learn to solve, with very little supervision, the decision variant of the Traveling Salesperson Problem (TSP), a highly relevant $\mathcal{NP}$-Complete problem. Our model is trained to function as an effective message-passing algorithm in which edges (embedded with their weights) communicate with vertices for a number of iterations after which the model is asked to decide whether a route with cost $<C$ exists. We show that such a network can be trained with sets of dual examples: given the optimal tour cost $C^{*}$, we produce one decision instance with target cost $x\%$ smaller and one with target cost $x\%$ larger than $C^{*}$. We were able to obtain $80\%$ accuracy training with $-2\%,+2\%$ deviations, and the same trained model can generalize for more relaxed deviations with increasing performance. We also show that the model is capable of generalizing for larger problem sizes. Finally, we provide a method for predicting the optimal route cost within $2\%$ deviation from the ground truth. 
In summary, our work shows that Graph Neural Networks are powerful enough to solve $\mathcal{NP}$-Complete problems which combine symbolic and numeric data.
\end{abstract} 

\section{Introduction}

    Deep learning has accomplished much in the last decade, advancing the state-of-the art of areas such as image recognition \cite{krizhevsky2012imagenet,simonyan2014very,li2015convolutional}, natural language processing \cite{cho2014learning,cho2014properties,bahdanau2014neural} and reinforcement learning \cite{mnih2013playing,mnih2015human,silver2016mastering,silver2017mastering}, which has been successfully combined with deep neural networks to master classic Atari games and yield superhuman performance in the Chinese boardgame Go \cite{mnih2013playing,mnih2015human,silver2016mastering,silver2017mastering}. However, the application of deep learning to symbolic domains directly, as opposed to their use in reinforcement learning agents, is still incipient \cite{GarcezLG2009,garcez2015neural,Evans_18}. 
    
    A promising meta-architecture to engineer models that learn on symbolic domains is to instantiate neural modules and assemble them in various configurations, each manifesting a graph representation of a given instance of the problem at hand \cite{scarselli2009graph}. In this context, the neural components can be trained to learn to compute messages to send between nodes, yielding a differentiable message-passing algorithm whose parameters can be improved via gradient descent. This technique has been successfully applied to a growing range of problem domains, although with different names. Gilmer et al., which apply it to quantum chemistry problems, adopt the term ``neural message passing'' \cite{gilmer2017neural}, while Palm et al. refer to ``recurrent relational networks'' in an attempt to train neural networks to solve Sudoku puzzles \cite{palm2017recurrent}. 
    
    A recent review of related techniques chooses the term \emph{graph networks} \cite{battaglia2018relational}, but we shall refer to \emph{graph neural networks}  named by Scarselli et al. who were among the first to propose such a model \cite{scarselli2009graph}. Graph Neural Networks (GNNs) have recently been successfully applied to the problem of predicting the boolean satisfiability of a CNF formula, a very relevant $\mathcal{NP}$-Complete combinatorial problem (SAT) \cite{selsam2018learning}. Selsam et al. show that GNNs can be trained to obtain satisfactory accuracy (approximately $85\%$) on small instances, and further that their performance can be improved by running the model for more message-passing timesteps. In addition, they show that satisfying assignments can be extracted from the network, which is never trained explicitly to produce them. The promising results of \emph{NeuroSAT} (as the authors named it) is an invitation to assess whether other hard combinatorial problems lend themselves to a simple GNN solution. 
    
    In this paper, we investigate whether GNNs can be trained to solve another $\mathcal{NP}$-Complete problem: the decision variant of the Traveling Salesperson Problem (TSP), assigned with deciding whether a given graph admits a Hamiltonian route with cost no greater than $C$. The NeuroSAT experiment from \cite{selsam2018learning} shows that graph neural networks can be trained to compute hard combinatorial problems, albeit for small instances. Nevertheless, SAT is a conceptually simpler problem, which can be defined purely in terms of boolean formulas. Thus, an open research question is to investigate whether GNNs can be trained to solve $\mathcal{NP}$-Complete problems involving numerical information (edge weights) in addition to symbolic relationships (edges or connections). The traveling salesperson problem in its decision variant (does graph $G$ admit a Hamiltonian path with cost $<C$?) is a promising candidate, as it requires edge weights $w_i$ as well as the ``target cost'' $C$ to be taken under consideration to compute a solution. 

The remainder of the paper is structured as follows. Next, we introduce a Graph Neural Network that shall be used in our TSP modelling. We then show how the proposed model learns to solve the Decision TSP and describe the experiments which validate the proposed model. Finally, we analyse the results and point out further research directions.

\section{A GNN Model for the Decision TSP}
Graph neural networks assign a multidimensional embedding $\in \mathbb{R}^d$ to each vertex in the graph representation of the problem instance at hand and perform a given number of message-passing iterations -- in which a neural module computes a message from each embedding and sends it along its adjacencies. Each vertex accumulates its incoming messages by adding them up (or aggregating them through any other operation) and feeding the resulting $\mathbb{R}^d$ vector into a Recurrent Neural Network (RNN) assigned with updating the embedding of said vertex. The only trainable parameters of such a model are the message computing modules and the RNN, so that conceptually what we have is a message-passing algorithm in which messages and updates are computed by neural networks.

Given a TSP instance $X = (\mathcal{G},C)$ composed of a graph $\mathcal{G} = (\mathcal{V}, \mathcal{E})$ and a target cost $C \in \mathbb{R}$, we could assign an embedding to each vertex and send messages alongside edges, but all information about edge weights would be lost this way. Instead, we additionally assign embeddings to \emph{edges}, which can be fed with their corresponding weights (edge embeddings in GNNs have shown promise in many applications \cite{battaglia2018relational}). In this context, we replace the vertex-to-vertex adjacency matrix $\mathbf{A} \in \{0,1\}^{|\mathcal{V}| \times |\mathcal{V}|}$, by an edge-to-vertex adjacency matrix $\mathbf{EV} \in \{0,1\}^{|\mathcal{E}| \times |\mathcal{V}|}$, which connects each edge $e_i = (s,t,w)$ to its source and target vertices. Because the model also needs to know the value of the target cost, we decided to feed $C$ to each edge embedding alongside with its corresponding weight: given a target cost $C$, for each edge $e_i = (s,t,w)$ we concatenate $w$ and $C$ to obtain a 2d vector $\in \mathbb{R}^2$. This vector is fed into a Multilayer perceptron (MLP) which expands it into $\mathbf{E}^{(1)}[i] \in \mathbb{R}^d$, the initial embedding for edge $e_i$. Following this initialization, the model undergoes a given number of iterations in which vertices and edges exchange messages and refine their embeddings, until finally the refined edge embeddings are fed into an MLP which computes a logit probability corresponding to the model's prediction of the answer to the decision problem. In summary, upon training our proposed model learns seven tasks:

\begin{enumerate}
    \item To produce a single $\mathbb{R}^d$ vector, which will be used to initialize all vertex embeddings
    \item A function $E_{init} : \mathbb{R}^2 \rightarrow \mathbb{R}^d$ to compute an initial edge embedding given the edge weight $w$ and the route cost $C$ (MLP)
    \item A function $V_{msg}: \mathbb{R}^d \rightarrow \mathbb{R}^d$ to compute a message to send to edges given a vertex embedding (MLP)
    \item A function $E_{msg}: \mathbb{R}^d \rightarrow \mathbb{R}^d$ to compute a message to send to vertices given an edge embedding (MLP)
    \item A function $V_{u}: \mathbb{R}^{2d} \rightarrow \mathbb{R}^{2d}$ to compute an updated vertex embedding (plus an updated RNN hidden state) given the current RNN hidden state and a message
    \item A function $E_{u}: \mathbb{R}^{2d} \rightarrow \mathbb{R}^{2d}$ to compute an updated edge embedding (plus an updated RNN hidden state) given the current RNN hidden state and a message
    \item A function $E_{vote}: \mathbb{R}^{d} \rightarrow \mathbb{R}^1$ to compute a logit probability given an edge embedding (MLP)
\end{enumerate}
Algorithm~\ref{alg:GNN-TSP} briefly summarizes the proposed GNN-based procedure to solve the decision TSP. In the sequel, we shall illustrate how the model is used in learning route costs and validate our architecture.

\begin{minipage}{\dimexpr\linewidth+0em}
\begin{algorithm}[H]
\caption{Graph Neural Network TSP Solver}\label{alg:GNN-TSP}
\begin{algorithmic}[1]
\Procedure{GNN-TSP}{$\mathcal{G} = (\mathcal{V},\mathcal{E}), C$}

\State
\LineComment{{\small Compute binary adjacency matrix from edges to source \& target vertices}}
\State $\mathbf{EV}[i,j] \negthickspace \leftarrow \negthickspace 1 \textrm{ iff } (\exists v' | e_i \negthickspace = \negthickspace (v_j,v',w)) |~ \forall e_i \negmedspace \in \negmedspace \mathcal{E}, v_j \negmedspace \in \negmedspace \mathcal{V}$

\State
\LineComment{{\small Compute initial edge embeddings}}
\State $\overset{(1)}{\mathbf{E}}[i] \leftarrow E_{init}(w,C) ~|~ \forall e_i = (s, t, w) \in \mathcal{E}$ 

\State
\LineComment{Run $t_{max}$ message-passing iterations}
\For{$t=1 \dots t_{max}$}
  \LineComment{{\small Refine each vertex embedding with messages received from edges in which it appears either as a source target vertex}}
  \label{alg:line:vertices_refinement}\State $\overset{(t+1)}{\mathbf{V}_h}, \overset{(t+1)}{\mathbf{V}} \negthickspace \leftarrow \negthickspace V_u(\overset{(t)}{\mathbf{V}_h}, \mathbf{EV}^{T} \negthickspace \times \negthickspace \underset{msg}{E}(\overset{(t)}{\mathbf{E}}))$
  \LineComment{{\small Refine each edge embedding with messages received from its source and its target vertex}}
  \label{alg:line:edges_refinement}\State $\overset{(t+1)}{\mathbf{E}_h}, \overset{(t+1)}{\mathbf{E}} \negthickspace \leftarrow \negthickspace E_u(\overset{(t)}{\mathbf{E}_h},\mathbf{EV} \negthickspace \times \negthickspace \underset{msg}{V}(\overset{(t)}{\mathbf{V}}))$
\EndFor

\LineComment{{\small Translate edge embeddings into logit probabilities}}
\State $\mathbf{E_{logits}} \leftarrow E_{vote}\left(\overset{t_{max}}{\mathbf{E}}\right)$
\LineComment{{\small Average logits and translate to probability (the operator $\langle \rangle$ indicates arithmetic mean)}}
\State $\textrm{prediction} \leftarrow \textrm{sigmoid}(\langle \mathbf{E_{logits}} \rangle)$

\EndProcedure
\end{algorithmic}
\end{algorithm}
\end{minipage}

\section{Training the Model}
\label{sub:training}
In order to train the GNN model, one has to provide it with four inputs: matrices $\mathbf{S}, \mathbf{T} \in \{0,1\}^{|\mathcal{E}| \times |\mathcal{V}|}$, the edge weights $\mathbf{D}$, a target cost $C \in \mathbb{R}$; the model is then trained with Stochastic Gradient Descent (SGD), more specifically by using TensorFlow's Adam \cite{kingma2014adam} implementation, to minimize the binary cross entropy loss between its prediction and the ground-truth (a boolean value indicating whether the answer to the decision problem is YES or NO).

To speed up training, it is convenient to perform SGD on batches with multiple instances. This can be achieved by performing the disjoint union between all graphs in the batch, yielding a ``batch'' graph with $n$ disjoint subgraphs. Because subgraphs are disjoint, messages will not traverse through any pair of them, and there will be no change to the embedding refinement process as compared to a single run. There will be logit probabilities computed for each edge in the batch graph, which can be averaged among individual instances to compute a prediction for each one of them. The binary cross entropy can then be computed between these predictions and the corresponding decision problem solutions.

We produce training instances by sampling $n \sim \mathcal{U}(20,40)$ random points on a $\frac{\sqrt{2}}{2} \times \frac{\sqrt{2}}{2}$ square and filling a distance matrix $\mathbf{D} \in \mathbb{R}^{n \times n}$ with the euclidean distance computed between each pair of points. These distances, by construction, are $\in [0,1]$. We also produce a complete adjacency matrix $\mathbf{A} \in \{0,1\}^{n \times n}$, and solve the corresponding TSP problem using the Concorde TSP solver \cite{hahsler2007tsp} to obtain optimal tour costs. A total of $2^{20}$ such graphs were produced, from which we sample a total of $1024$ per epoch to ensure that the probability of the model seeing the same graph twice at training time is kept low. Finally, for each graph $\mathcal{G}$ with optimal tour cost $C^{*}$ we produce two decision instances $X^{+} = (\mathcal{G},1.02 C^{*})$ and $X^{-} = (\mathcal{G},0.98 C^{*})$ for which the answers are by construction YES and NO respectively. In doing so we effectively train the model to predict the decision problem within a $2\%$ positive or negative deviation from the optimal tour cost. 

The model is instantiated with $64$-dimensional embeddings for vertices and edges and three-layered (64,64,64) MLPs with ReLU nonlinearities as the activations for all layers except for the last one, which has a linear activation. The model is run for $T_{max} = 32$ time steps of message-passing.


\section{Experimental Results and Analyses}

\begin{figure}[h]
  \centering
  \includegraphics[width=\linewidth]{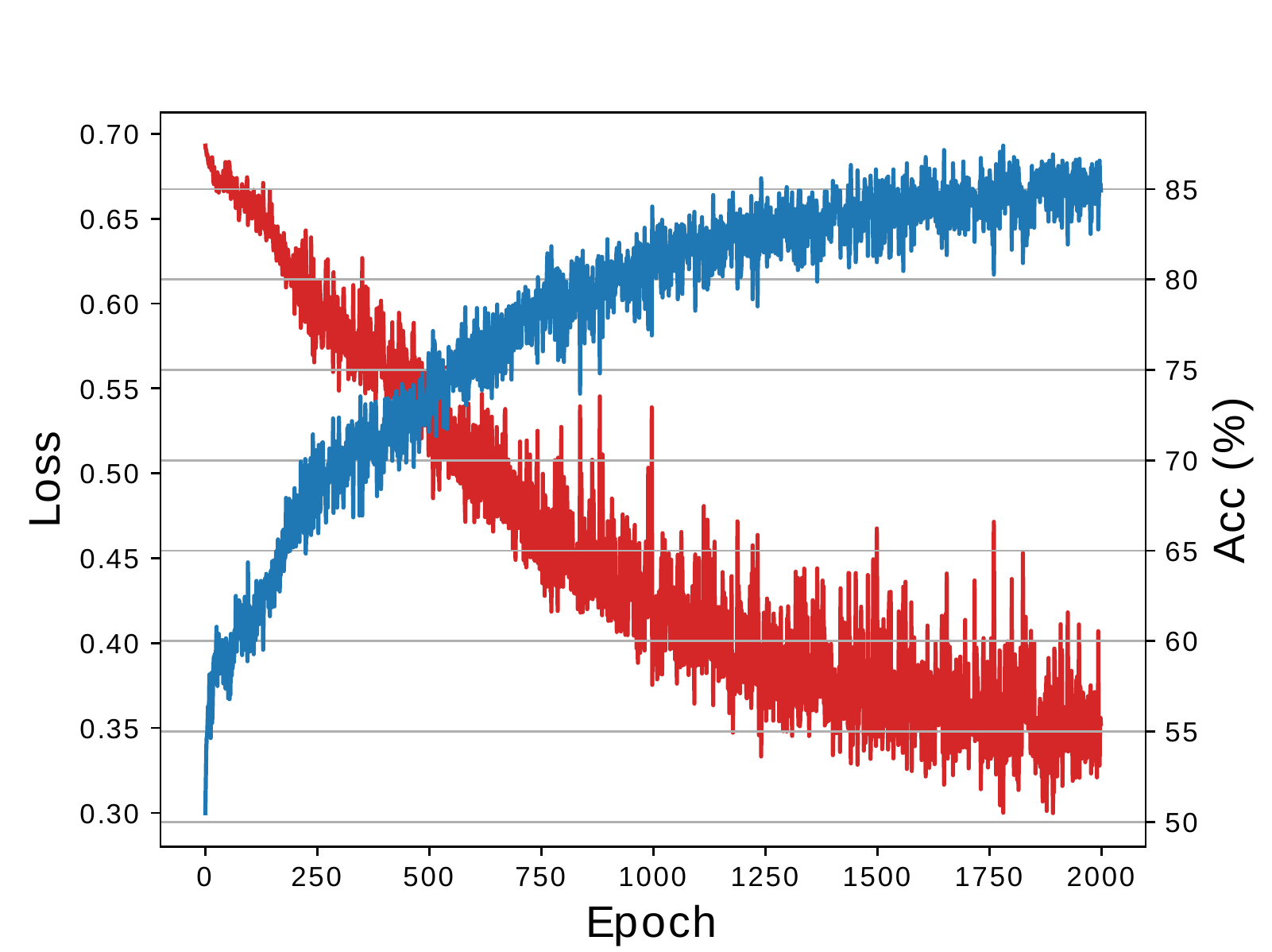}
  \caption{Evolution of the binary cross entropy loss (downward curve in red) and accuracy (upward curve in blue) throughout a total of $2000$ training epochs on a dataset of $2^{20}$ graphs with $n \sim \mathcal{U}(20,40)$. Each graph with optimal TSP route cost $C^{*}$ is used to produce two instances to the TSP decision problem -- ``is there a route with cost $<1.02 C^{*}$?'' and ``is there a route with cost $<0.98 C^{*}$?'', which are to be answered with YES and NO respectively. Each epoch is composed of $128$ batches of $16$ instances each (please note that at each epoch the network sees only a small sample of the dataset, and the accuracy here is computed relative to it).}
  \label{fig:training-decision}
\end{figure}

\begin{figure}[h]
  \centering
  \includegraphics[width=\linewidth]{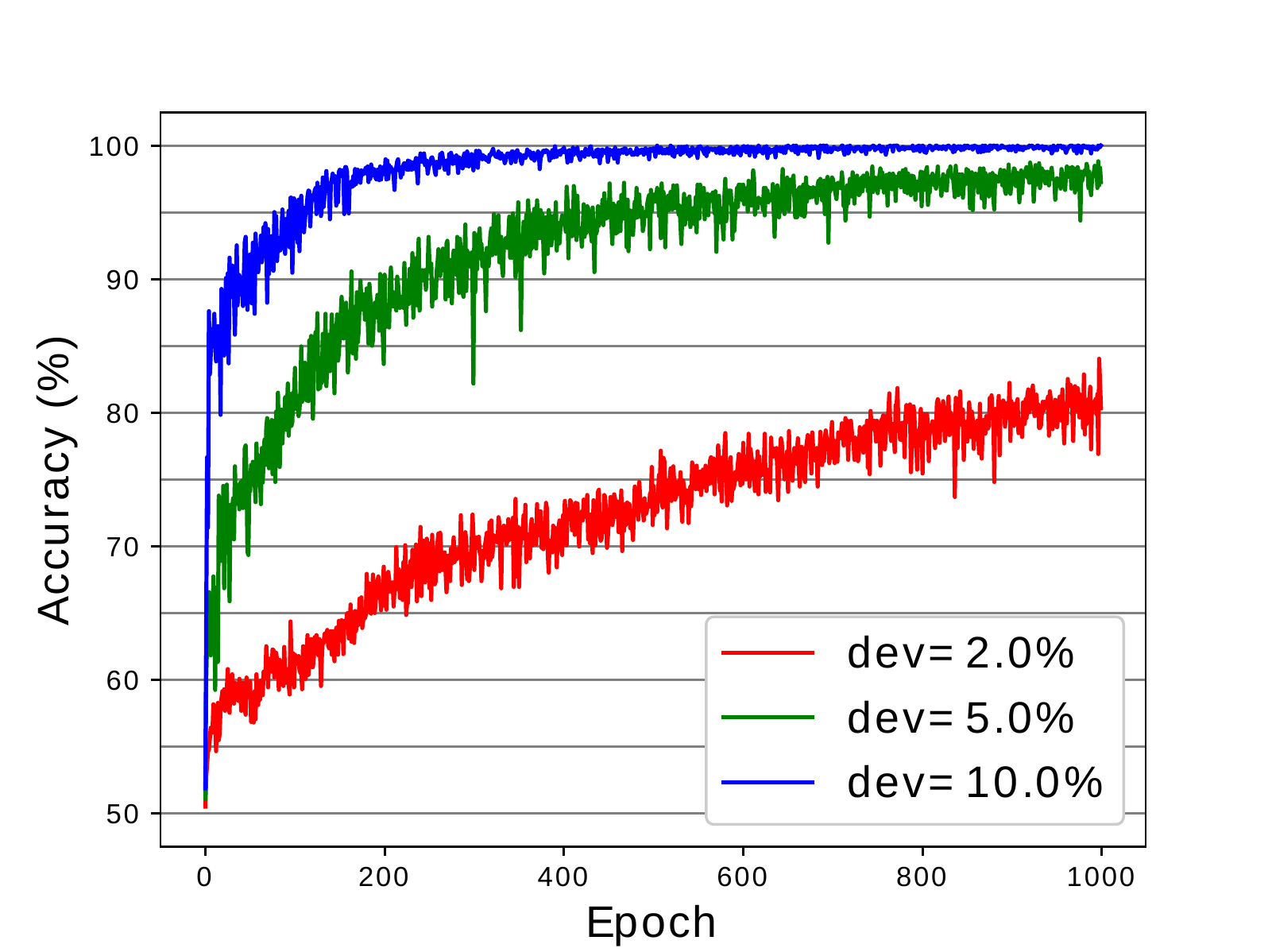}
  \caption{The larger the deviation from the optimal cost, the faster the model learns: we were able to obtain $>95\%$ accuracy for $10\%$ deviation in 200 epochs. For $5\%$, that performance requires double the time. For $2\%$ deviation, two thousand epochs are required to achieve $85\%$ accuracy.}
  \label{fig:compare-trainings}
\end{figure}

Upon $2000$ training epochs, the model achieved $80.16\%$ accuracy averaged over the $2^{21}$ instances of the training set, having also obtained $80\%$ accuracy on a testing set of $2048$ instances it had never seen before. Instances from training and test datasets were produced with the same configuration ($n \sim \mathcal{U}(20,40)$ and $2\%$ percentage  deviation). Figure~\ref{fig:training-decision} shows the evolution of the binary cross entropy loss and accuracy throughout the training process. Note that it is much easier to train the model with more relaxed deviations from the optimal cost, as Figure~\ref{fig:compare-trainings} shows.
 
\subsection{Extracting Route Costs}
\label{ssec:extracting-route-costs}

\begin{figure}[h]
    \centering
    \includegraphics[width=\linewidth]{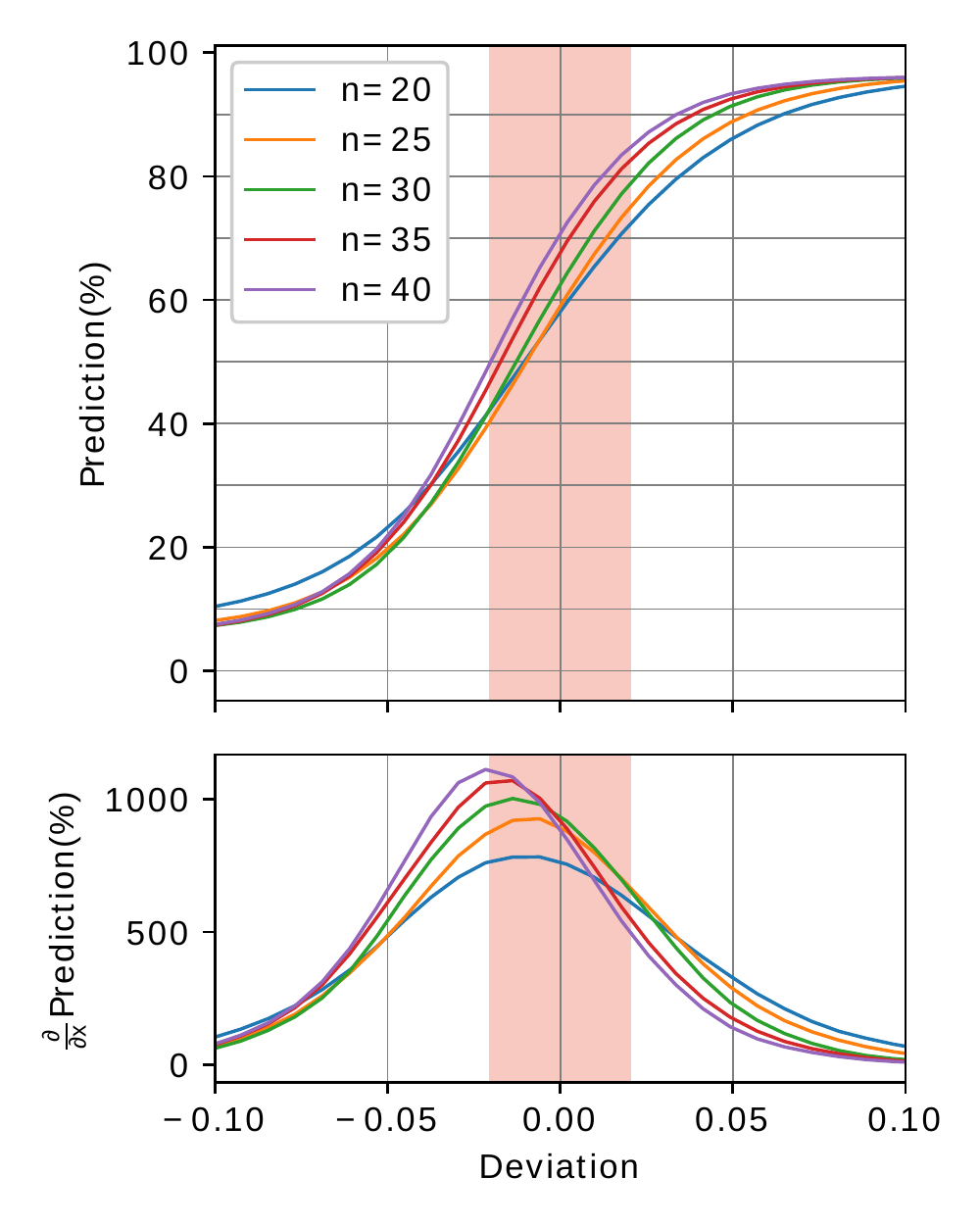}
    \caption{Average prediction obtained from the model as a function of the deviation between the target cost and the optimal cost for varying instance sizes (the pink band indicates the $[-2\%,+2\%]$ interval). As expected, the curve is S-shaped, signalling that the model is very confident that routes with sufficiently large/small costs do/do not exist. The average prediction undergoes a phase transition as we traverse from negative to positive deviations. Larger instances exhibit smaller critical points, as evidenced by the left shifts on the derivatives of the acceptance curves in the bottom subfigure. The prediction for each deviation is averaged over $1024$ instances.}
    \label{fig:acceptance-curves}
\end{figure}

Now that we have obtained a solver for the decision TSP, we can exploit it to yield route cost predictions within a reasonable margin from the optimal cost. Figure~\ref{fig:acceptance-curves} shows how the model behaves when it is asked to solve the decision problem for varying target costs. From its characteristic S-shape we can learn that the model feels confident that routes with too small costs do not exist and also confident that routes with too large costs do exist. Between these two regimes, the prediction undergoes a phase transition, with the model becoming increasingly unsure as we approach zero deviation from the optimal cost. In fact, this ``acceptance curve'' plotted for varying instance sizes is reminiscent of phase transitions on hard combinatorial problems such as SAT \cite{dudek2017combining} and, along with a large number of $\mathcal{NP}$-Hard problems, the TSP itself has been shown to exhibit phase transition phenomena \cite{kirkpatrick1985configuration,zhang2004phase}. 

More importantly, we know from theoretical results that the average TSP tour length for a set of $n$ random (uniform) points on a plane is asymptotically proportional to $\sqrt{n}$ with the two-dimensional ``TSP constant'' $\beta(2)$ as a proportionality factor \cite{beardwood1959shortest}. As a corollary, large instances allow for proportionally shorter routes than small instances\footnote{$\displaystyle\lim_{n \to \infty}{ C_n^{*} / n} = \lim_{n \to \infty}{ \beta(2) \sqrt{n} / n} = 0$ where $C_n^{*}$ is the optimal tour cost for a $n$-city instance}, a fact that, we believe, is manifest in the curves of Figure~\ref{fig:acceptance-curves}: for deviations close to zero, the model feels more confident that a route exists the larger the instance size is. As a result, the critical point (the deviation at which the model starts guessing YES) undergoes a left shift as the instance size increases, as seen in the curves' derivatives in Figure~\ref{fig:acceptance-curves}.

In addition, all acceptance curves are above the $50\%$ line for deviation $=0$, from which we conjecture that the trained model guesses by default that a route \textbf{does} exist and proceeds to \textbf{disprove} this claim throughout message-passing iterations. Interestingly, this behavior is opposite to that of the GNN SAT-solver \emph{NeuroSAT} \cite{selsam2018learning}, which guesses UNSAT by default and changes its prediction only upon finding a satisfiable assignment. The factors determining which strategy the model will learn remain an open question, but we are hopeful that it is possible to engineer a training set to enforce that the model learns a negative-by-default algorithm.

To the best of our knowledge, the curves in Figure~\ref{fig:acceptance-curves} become arbitrarily close to zero as we progress towards smaller deviations, but unfortunately the model starts to lose confidence that a route exists when it is fed with large target costs ($\approx 100\%$ deviation). This is probably due to the fact that, being trained with $-2\%,+2\%$ deviations, the model has never seen target costs that large. Fortunately this can be corrected by re-training it for a single epoch with $-2\%, +2\%, +100\%, +200\%, +1000\%$ deviations, which is done with no significant effect to the test accuracy.

Intuitively, if we know nothing about the optimal cost, we can assume that we are closest to its value when the model's predictions are closest to $50\%$. We can therefore guess an initial cost and perform a binary search on the $x$-axis of Figure~\ref{fig:acceptance-curves}. The procedure is detailed in Algorithm~\ref{alg:binary-search}.

\begin{algorithm}[h]
\caption{Binary Search}\label{alg:binary-search}
\begin{algorithmic}[1]
\Procedure{Binary-Search}{$\mathcal{G} = (\mathcal{V},\mathcal{E})$, $p$, $\delta$}

\LineComment{Choose an initial guess for the optimal route cost. $w^{n-}$ and $w^{n+}$ are the sets of the costs of the $n$ edges $\in \mathcal{E}$ with smallest / largest costs respectively.}
\State $C_{min} \leftarrow \sum{w^{n-}_i}$
\State $C_{max} \leftarrow \sum{w^{n+}_i}$
\State $C \sim \mathcal{U}(C_{min},C_{max})$
\While{$C_{min} < C (1-\delta) \vee C (1+\delta) < C_{max}$}
    \If{$\textrm{GNN-TSP}(\mathcal{G},C) < p$}
        \State $C_{min} \leftarrow C$
    \Else
        \State $C_{max} \leftarrow C$
    \EndIf
    \State $C \leftarrow (C_{min}+C_{max})/2$
\EndWhile

\Return $C$

\EndProcedure
\end{algorithmic}
\end{algorithm}

Instantiated with $\delta = 0.01$ and using the weights after the training and the single epoch of training for greater deviations, Algorithm~\ref{alg:binary-search} is able to predict route costs with on average $1.5\%$ absolute deviation from the optimal, running for on average $8.9$ iterations on the test dataset ($1024$ n-city graphs with $n \sim \mathcal{U}(20,40)$). 

\subsection{Model Performance on Larger Instances}

\begin{figure}[h]
    \centering
    \includegraphics[width=\linewidth]{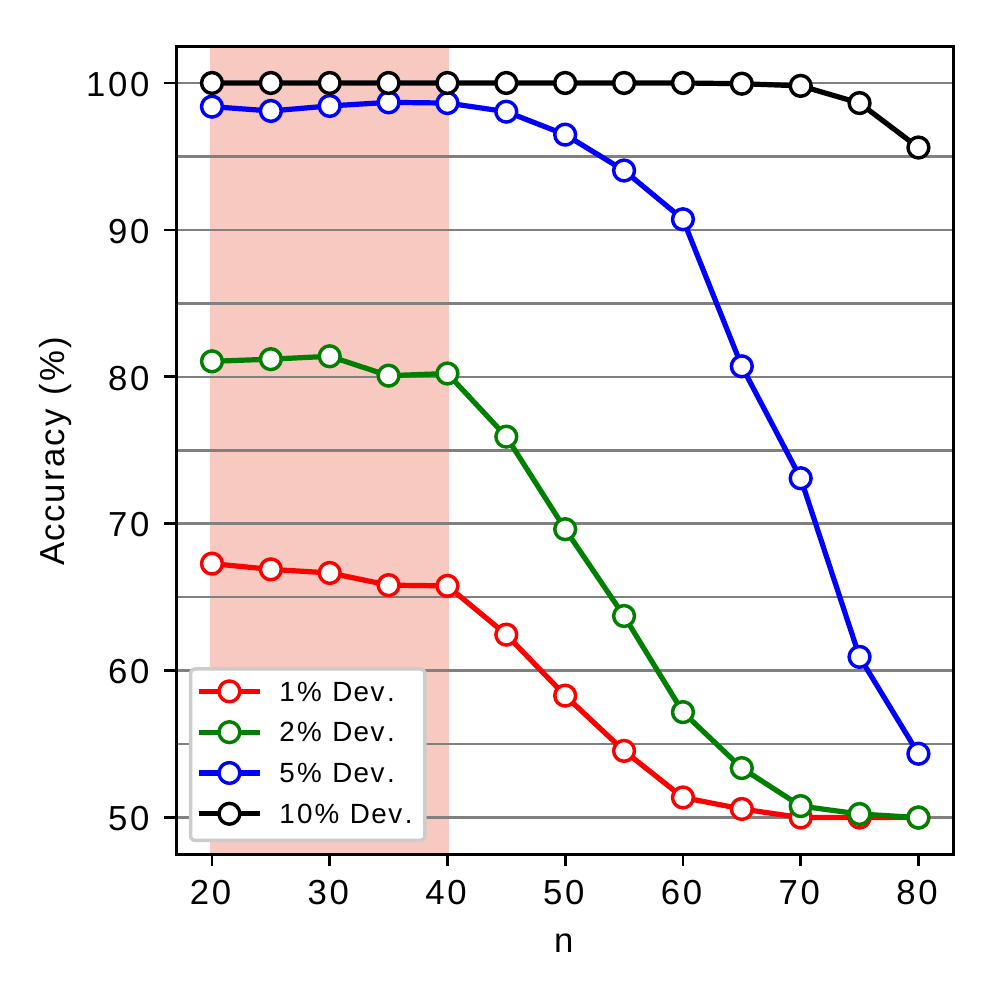}
    \caption{Accuracy of the trained model evaluated on datasets of $1024$ instances with varying numbers of cities ($n$). The model is able to obtain $>80\%$ accuracy for $-2\%,+2\%$ deviation on the range of sizes it was trained on (painted in pink), but its performance degenerates progressively for larger instance sizes before reaching the baseline of $50\%$ at $n \approx 75$. Larger deviations yield higher accuracy curves, with the model obtaining $>95\%$ accuracy for $-10\%,+10\%$ deviation even for the largest instance sizes.}
    \label{fig:test-varying-sizes}
\end{figure}

The model was trained on instances with no more than $n=40$ cities, but we wanted to know to what extent the learned algorithm generalizes to larger problem sizes. We averaged the trained model accuracy over test datasets of $1024$ instances for varying values of $n$, for which the results are shown in Figure~\ref{fig:test-varying-sizes}. We found that the model is able to sustain $>80\%$ accuracy throughout the range of sizes it was trained on, but loses performance progressively for larger problem sizes until it reaches the baseline of $50\%$. Also, as expected given the acceptance curves in Figure~\ref{fig:acceptance-curves}, the model performs better for larger deviations ($5\%$, $10\%$) and worse for smaller ones ($1\%$). Do note as well that a problem of double the size would require $2^{n}$ more time to compute by traditional algorithms, and thus such a rapid decay in accuracy is to be expected.

\subsection{Generalizing to Larger Deviations}
\begin{figure}[h]
    \centering
    \includegraphics[width=1\linewidth]{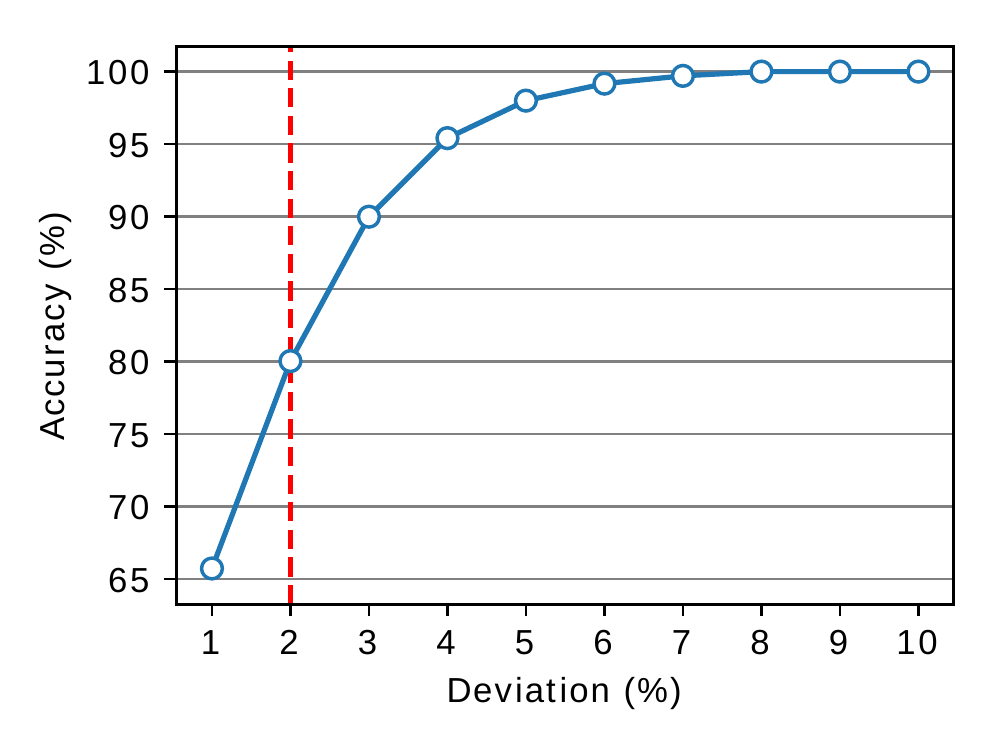}
    \caption{Accuracy of the trained model evaluated on the same test dataset of $1024$ n-city instances with $n \sim \mathcal{U}(20,40)$ for varying deviations from the optimal tour cost. Although it was trained with target costs $-2\%,+2\%$ from the optimal (dashed line), the model can generalize for larger deviations with increasing accuracy. Additionally, it could still obtain accuracies above the baseline ($50\%$) for instances more constrained than those it was trained on, with $65\%$ accuracy at $-1\%,+1\%$.}
    \label{fig:test-varying-dev}
\end{figure}

Both the acceptance curves in Figure~\ref{fig:acceptance-curves} and the accuracy curves in Figure~\ref{fig:test-varying-sizes} suggest that the model generalizes to larger deviations from the optimal tour cost than the $2\%$ it was trained on. In fact, these curves suggest that the model becomes more confident the larger the deviation is, which is not surprising given that the corresponding decision instances are comparatively more relaxed. Figure~\ref{fig:test-varying-dev} shows how the accuracy increases until it plateaus at $\approx100\%$ for increasing deviations and Table~\ref{tab:acc-by-dev} 
depicts these results in the validation test sets.

\begin{table}[h]
    \centering
    \begin{tabular}{cc}
         \toprule
         Deviation & Accuracy (\%) \\ \midrule
          $1$ & $66$ \\
          $2$ & $80$ \\
          $5$ & $98$ \\
         $10$ & $100$ \\ \bottomrule
    \end{tabular}
    \caption{Test accuracy averaged over $1024$ n-city instances with $n \sim \mathcal{U}(20,40)$ for varying percentage deviations from the optimal route cost.}
    \label{tab:acc-by-dev}
\end{table}

\subsection{Baseline Comparison}

We chose to train the model on decision instances with $-2\%, +2\%$ deviation from the optimal tour cost not because this was our intended performance, but because $2\%$ was the smallest deviation for which the network could be trained within reasonable time ($\leq 2000$ epochs). For this reason, we do not know initially how the trained model compares with other methods. Although our goal is not to produce a state-of-the-art TSP solver but rather to demonstrate that neural networks can learn to solve this problem with very little supervision (two bits: one bit for a positive solution and one bit for a negative one), we want to evaluate whether our model can outperform simple heuristics. We compare our model with (1) a Nearest Neighbor (NN) route construction and (2) a Simulated Annealing (SA) routine \cite{kirkpatrick1983optimization}. NN is arguably the simplest TSP heuristic, generally yielding low quality solutions. SA can generally produce good routes for the euclidean TSP, if the meta-parameters are calibrated correctly. We calibrate the SA's initial temperature $T$, cooling rate $\alpha$ and stopping temperature $T_{min}$ with the \emph{irace} automatic algorithm configuration package \cite{irace}.

\begin{figure}[h]
    \centering
    \includegraphics[width=\linewidth]{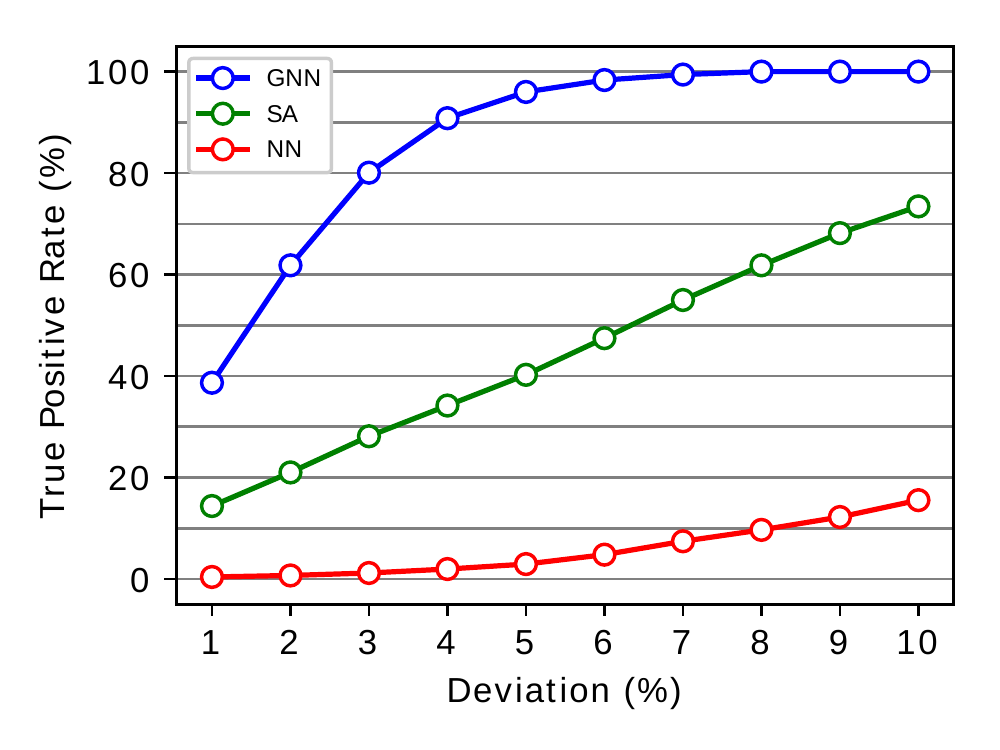}
    \caption{Nearest Neighbor (NN) and Simulated Annealing (SA) do not yield a prediction for the decision variant of the TSP but rather a feasible route. To compare their performance with our model's, we evaluate the frequency in which they yield solutions below a given deviation from the optimal route cost and plot alongside with the True Positive Rate (TPR) of our model for the same test instances ($1024$ n-city graphs with $n \sim \mathcal{U}(20,40)$).}
    \label{fig:test-varying-dev-baseline}
\end{figure}

Figure~\ref{fig:test-varying-dev-baseline} compares the True Positive Rate (TPR) of the trained model with the frequency in which these two heuristics could produce routes within a given deviation from the optimal route cost. This frequency can be thought as the TPR obtained by converting these methods into a predictor for the decision variant of the same problem (guess YES whenever you can constructively prove that a route within the target cost exists and NO otherwise). For the test dataset ($1024$ n-city graphs with $n \sim \mathcal{U}(20,40)$), Nearest Neighbor obtains on average routes $20.2\%$ more expensive than the optimal, while Simulated Annealing brings that number down to $6.7\%$. Nevertheless, for all tested deviations, the trained GNN model outperforms both methods, obtaining $>90\%$ TPR from deviations $4\%$ and above.

\subsection{Generalizing to Other Distributions}

Although the model was trained on two-dimensional euclidean graphs, it can generalize, to some extent, to more comprehensive distributions. To evaluate this, we considered two families of graphs obtained from uniformly random distance matrices: in the first distribution (``Rand.'' in Table~\ref{tab:test_other_distr}), edge weights are simply sampled uniformly at random; in the second (``Rand. Metric'' in Table~\ref{tab:test_other_distr}), edge weights are first sampled uniformly at random and then the metric property is enforced by replacing edge weights by the shortest path distance between the corresponding vertices. For $2\%$ deviation from the optimal tour cost, the model was able to obtain $64\%$ accuracy on the random metric instances (versus $80\%$ on euclidean), but the performance is better for more relaxed deviations, with $82\%$ at $5\%$ and $96\%$ at $10\%$ deviation from the optimal route cost. The model was unable to achieve performance above the $50\%$ baseline for non-metric instances. We also evaluated the model with real world instances gathered from the Tsplib95 dataset \cite{tsplib95}, for which the results obtained for the trained model with Algorithm~\ref{alg:binary-search} are reported in Table~\ref{tab:test-real-world}. In general, the model underestimates the optimal route cost, which is expected given the discussion in subsection on \emph{Extracting route costs} above. When the absolute relative deviation is considered, the GNN outperforms the SA routine for 6 out of 9 instances.

\begin{table}[h]
    \centering
    \begin{tabular}{cccc}
         \toprule
         Deviation & \multicolumn{3}{c}{Accuracy (\%)}  \\ 
              & Euc. 2D & Rand. Metric & Rand. \\ \midrule
          $1$ &    $66$ &         $57$ & $50$  \\
          $2$ &    $80$ &         $64$ & $50$  \\
          $5$ &    $98$ &         $82$ & $50$  \\
         $10$ &   $100$ &         $96$ & $50$  \\ \bottomrule
    \end{tabular}
    \caption{Test accuracy averaged over $1024$ n-city instances with $n \sim \mathcal{U}(20,40)$ for varying percentage deviations from the optimal route cost for differing random graph distributions: two-dimensional euclidean distances, ``random metric'' distances and random distances.}
    \label{tab:test_other_distr}
\end{table}

\begin{table}[]
\centering
\begin{threeparttable}
\begin{tabular}{ccrr}
\toprule
          Instance & Size & \multicolumn{2}{c}{Relative Deviation (\%)} \\
          &      &      GNN &       SA \\ \midrule
          \rowcolor{red!50!white!50}
          ulysses16\tnote{1} &   16 & $-22.80$ & $ +1.94$ \\
          \rowcolor{red!50!white!50}
          ulysses22\tnote{1} &   22 & $-27.20$ & $ +1.91$ \\
          \rowcolor{red!50!white!50}
          eil51              &   51 & $-18.37$ & $+18.07$ \\
          \rowcolor{blue!50!white!50}
          berlin52           &   52 & $ -8.73$ & $+21.45$ \\
          \rowcolor{blue!50!white!50}
          st70               &   70 & $-11.87$ & $+14.47$ \\
          \rowcolor{blue!50!white!50}
          eil76              &   76 & $-13.91$ & $+19.24$ \\
          \rowcolor{blue!50!white!50}
          kroA100            &  100 & $ -2.00$ & $+30.73$ \\
          \rowcolor{blue!50!white!50}
          eil101             &  101 & $ -9.93$ & $+20.46$ \\
          \rowcolor{blue!50!white!50}
          lin105             &  105 & $ +6.37$ & $+17.77$ \\ \bottomrule
\end{tabular}
\begin{tablenotes}
\item[1] \footnotesize{These instances had their distance matrix computed according to Haversine formula (great-circle distance).}
\end{tablenotes}
\end{threeparttable}
\caption{The relative deviations from the optimal route cost are compared for the prediction obtained from the trained model with Algorithm~\ref{alg:binary-search} (GNN) and the Simulated Annealing heuristic (SA). Lines referring to instances in which the trained model {\color{blue}outperformed} and {\color{red}underperformed} the SA heuristic are colored {\color{blue}blue} and {\color{red}red} respectively. Note that deviations obtained from the trained model are negative in general, as expected given the discussion in the subsection about  Extracting route costs above.}
\label{tab:test-real-world}
\end{table}

\subsection{Implementation and Reproducibility}
The reproducibility of machine learning studies and experiments is relevant to the field of AI given the myriad of parameters and implementation decisions one has to make. With this in mind, we summarize here the instantiation parameters of our model. The embedding size was chosen as $d=64$, all message-passing MLPs are three-layered with layer sizes $(64,64,64)$ with ReLU nonlinearities as the activation of all layers except the last one, which has a linear activation. The edge embedding initialization MLPs are three-layered with layer sizes ($8,16,32$) (we tried different architectures but have only obtained success with increasing layer sizes and a small initial layer). The kernel weights are initialized with TensorFlow's Xavier initialization method described in \cite{glorot2010understanding} and the biases are initialized with zeroes. The recurrent unit assigned with updating embeddings is a layer-norm LSTM \cite{ba2016layer} with ReLU as its activation and both with kernel weights and biases initialized with TensorFlow's Glorot Uniform Initializer \cite{glorot2010understanding}, with the addition that the forget gate bias were increased by 1. The number of message-passing timesteps is set at $t_{max} = 32$. For each graph instance, a pair of decision instances was created: a negative instance with target cost $2\%$ smaller than the optimal and a positive instance with target cost $2\%$ greater than the optimal. 
The training instances can be randomized but it is important that these pairs remain together in the same batch. Each training epoch is composed by $128$ Stochastic Gradient Descent operations on batches of $16$ instance pairs (with a positive and with a negative deviation) each, randomly sampled from the training dataset. 

Since we have the liberty of generating our own training instances and to mitigate overfitting effects, we produced $2^{20}$ n-city graphs with $n \sim \mathcal{U}(20,40)$. Instances can be batched together by performing a disjoint union on a set of $n$ graphs, producing a graph with $n$ connected components in which information flow does not ``spill'' from one to another. Finally, on all experiments we have normalized all edge weights to be $\in [0,1]$, and the target cost is always normalized by the number of cities $n$.
We have dedicated significant effort into making the reproduction of the experiments reported here available as a plug-and-play functionality. The code used to generate instances, train and evaluate the model and produce the figures presented in this paper is available at \url{https://github.com/machine-reasoning-ufrgs/TSP-GNN}.

\section{Conclusions and Future Work}

In this paper, we have proposed a Graph Neural Network (GNN) architecture which assigns multidimensional embeddings to vertices and edges in a graph. In our model, vertices and edges undergo a number of message-passing iterations in which their embeddings are enriched with local information. Finally, each embedding ``votes'' on whether the graph admits a Traveling Salesperson route no longer than $C$, and the votes are combined to yield a prediction. We show that such a network can be trained with sets of dual decision instances: given a optimal cost $C^{*}$, we produce a (negative) instance with target cost $x\%$ smaller and a (positive) instance with target cost $x\%$ larger than $C^{*}$. Upon training the model with $-2\%,+2\%$ deviations were able to obtain $80\%$ accuracy, and the model learned to generalize to larger deviations with increasing accuracy ($96\%$ at $-5\%,+5\%$). We also show how the model generalizes to some extent to larger problem sizes and different distributions. We conjecture that the model learns a positive-by-default algorithm, initially guessing that a route does exist and overriding that prediction when it can convince itself that is does not. In addition, the network is more confident that a route exists the larger the problem size is, which we think reflects the fact that the optimal TSP tour for a n-city euclidean graph scales with $\sqrt{n}$ (and therefore larger graphs admit proportionally shorter routes). By plotting the ``acceptance curves'' of the trained model, we uncovered a behavior reminiscent of phase transitions on hard combinatorial problems. Coupled with a binary search, these curves allow for an accurate prediction of the optimal TSP cost, even though the network was only trained to provide yes-or-no answers.

We are hopeful that a training set can be engineered in such a way as to enforce the model to learn a negative-by-default algorithm, possibly enabling us to extract a TSP route from the refined embeddings as we know to be possible given the NeuroSAT experiment \cite{selsam2018learning}. We intend on training and evaluating our model on a comprehensive set of real and random graphs, and to assess how far the model can generalize to larger problem sizes compared to those it was trained on. Finally, we believe that this experiment can showcase the potential of GNNs to the AI community and help promote an increased interest on integrated machine learning and reasoning models.
\section{Acknowledgments}
This research was partly supported by Coordenação de Aperfeiçoamento de Pessoal de Nível Superior (CAPES) - Finance Code 001 and by the Brazilian Research Council CNPq. 
\bibliographystyle{named}
\bibliography{bib}

\end{document}